\documentclass{article}

\usepackage[numbers]{natbib}
\usepackage[preprint]{neurips_2023}

\usepackage[dvipsnames]{xcolor}         
\definecolor{linkColor}{rgb}{0.18,0.39,0.62}
\usepackage[utf8]{inputenc} 
\usepackage[T1]{fontenc}    
\usepackage[colorlinks=true,linkcolor=linkColor,citecolor=linkColor,filecolor=linkColor,urlcolor=linkColor]{hyperref}       
\usepackage{url}            
\usepackage{booktabs}       
\usepackage{amsfonts}       
\usepackage{nicefrac}       
\usepackage{microtype}      

\usepackage{graphicx}
\usepackage{arydshln}

\usepackage{booktabs}
\usepackage{multirow}
\usepackage{caption}
\usepackage{subcaption}
\usepackage{makecell}
\usepackage{csquotes}
\usepackage{epigraph}

\RequirePackage{algorithm}
\RequirePackage{algorithmic}

\usepackage{multirow}
\usepackage{amsmath}
\usepackage{capt-of}
\usepackage{tabularx}
\usepackage{epsfig}
\usepackage{amssymb}
\usepackage{amsfonts}
\usepackage{booktabs}
\usepackage{scalerel}
\usepackage[inline]{enumitem}
\usepackage{listings}
\usepackage{varwidth}
\usepackage[export]{adjustbox}
\usepackage{tikz}
\usetikzlibrary{tikzmark}

\usepackage{stmaryrd}
\usepackage{bbm}
\usepackage{wrapfig}
\usepackage{pifont}
\usepackage[noabbrev]{cleveref}

\newcommand{\tabincell}[2]{\begin{tabular}{@{}#1@{}}#2\end{tabular}}

\definecolor{deepblue}{rgb}{0,0,0.5}
\definecolor{officeblue}{RGB}{0,102,204}
\definecolor{deepred}{rgb}{0.6,0,0}
\definecolor{deepgreen}{rgb}{0,0.5,0}
\definecolor{mybrickred}{RGB}{182,50,28}

\definecolor{fillcolor}{RGB}{216,217,252}

\usepackage{etoolbox}
\usepackage{framed}

\newif\ifxetexorluatex
\ifxetex
  \xetexorluatextrue
\else
  \ifluatex
    \xetexorluatextrue
  \else
    \xetexorluatexfalse
  \fi
\fi
\newcommand*\quotesize{60} 
\newcommand*{\openquote}
   {\tikz[remember picture,overlay,xshift=-4ex,yshift=-2.5ex]
   \node (OQ) {\fontsize{\quotesize}{\quotesize}\selectfont``};\kern0pt}

\newcommand*{\closequote}[1]
  {\tikz[remember picture,overlay,xshift=4ex,yshift={#1}]
   \node (CQ) {\fontsize{\quotesize}{\quotesize}\selectfont''};}

\colorlet{shadecolor}{white}

\newcommand*\shadedauthorformat{\emph} 

\newcommand*\authoralign[1]{%
  \if#1l
    \def\authorfill{}\def\quotefill{\hfill}
  \else
    \if#1r
      \def\authorfill{\hfill}\def\quotefill{}
    \else
      \if#1c
        \gdef\authorfill{\hfill}\def\quotefill{\hfill}
      \else\typeout{Invalid option}
      \fi
    \fi
  \fi}
\newenvironment{shadequote}[2][l]%
{\authoralign{#1}
\ifblank{#2}
   {\def\shadequoteauthor{}\def\yshift{-2ex}\def\quotefill{\hfill}}
   {\def\shadequoteauthor{\par\authorfill\shadedauthorformat{#2}}\def\yshift{2ex}}
\begin{snugshade}\begin{quote}\openquote}
{\shadequoteauthor\quotefill\closequote{\yshift}\end{quote}\end{snugshade}}

\newfloat{Code}{htbp}{Code}

\usepackage{amsmath,amsfonts,bm}

\def\eqref#1{equation~(\ref{#1})}
\def\Eqref#1{Equation~(\ref{#1})}

\def\1{\bm{1}}

\def\vk{{\bm{k}}}

\def\vq{{\bm{q}}}

\def\vs{{\bm{s}}}

\def\vw{{\bm{w}}}
\def\vx{{\bm{x}}}

\DeclareMathAlphabet{\mathsfit}{\encodingdefault}{\sfdefault}{m}{sl}
\SetMathAlphabet{\mathsfit}{bold}{\encodingdefault}{\sfdefault}{bx}{n}

\newcommand{\softmax}{\mathrm{softmax}}

\newcommand\our{RetNet}

\title{Retentive Network: A Successor to Transformer \\ for Large Language Models}

\author{
Yutao Sun\thanks{~Equal contribution. $\diamond$ Corresponding author.}$~~^{\dag\ddag}$~~~~Li Dong\footnotemark[1]$~~^{\dag}$~~~~Shaohan Huang$^{\dag}$~~~Shuming Ma$^{\dag}$ \\
\bf Yuqing Xia$^{\dag}$~~~Jilong Xue$^{\dag}$~~~Jianyong Wang$^{\ddag}$~~~Furu Wei$^{\dag}$$^{\diamond}$ \\
$^\dag$ Microsoft Research ~~~~
$^\ddag$ Tsinghua University \\
{\href{https://aka.ms/GeneralAI}{https://aka.ms/GeneralAI}}
}

\begin{document}

\maketitle

\begin{abstract}
In this work, we propose \textbf{Retentive Network} (\textsc{RetNet}) as a foundation architecture for large language models, simultaneously achieving training parallelism, low-cost inference, and good performance. We theoretically derive the connection between recurrence and attention. Then we propose the retention mechanism for sequence modeling, which supports three computation paradigms, i.e., parallel, recurrent, and chunkwise recurrent. Specifically, the parallel representation allows for training parallelism. The recurrent representation enables low-cost $O(1)$ inference, which improves decoding throughput, latency, and GPU memory without sacrificing performance. The chunkwise recurrent representation facilitates efficient long-sequence modeling with linear complexity, where each chunk is encoded parallelly while recurrently summarizing the chunks. Experimental results on language modeling show that \textsc{RetNet} achieves favorable scaling results, parallel training, low-cost deployment, and efficient inference. The intriguing properties make \textsc{RetNet} a strong successor to Transformer for large language models. Code will be available at \url{https://aka.ms/retnet}.
\end{abstract}


\vfill

\begin{figure*}[ht]
\centering
\centering
\captionsetup{type=figure}
\includegraphics[width=0.92\textwidth]{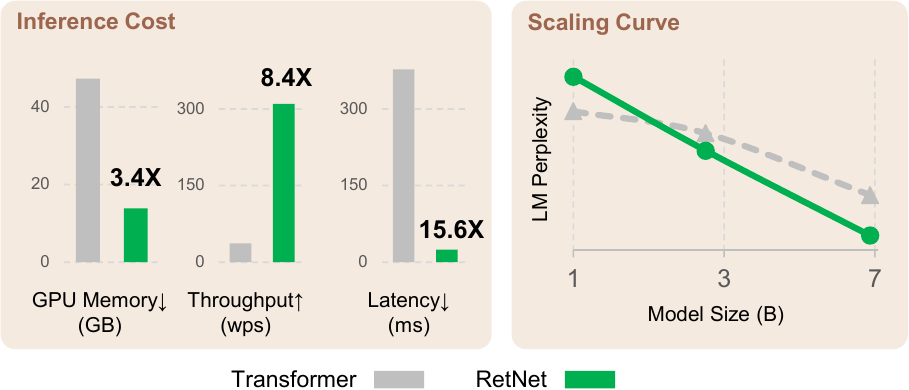}
\caption{Retentive network (\our{}) achieves low-cost inference (i.e., GPU memory, throughput, and latency), training parallelism, and favorable scaling curves compared with Transformer. Results of inference cost are reported with 8k as input length. Figure~\ref{fig:infer:cost} shows more results on different sequence lengths.}
\label{fig:teaser}
\end{figure*}

\vfill

\clearpage

\begin{shadequote}[r]{\small Arthur C. Clarke}
{\small The only way to discover the limits of the possible is to go beyond them into the impossible.}
\end{shadequote}

\section{Introduction}
\label{sec:intro}

\begin{wrapfigure}{r}{0.5\textwidth}
\vspace{-20pt}
\begin{center}
\includegraphics[width=0.48\textwidth]{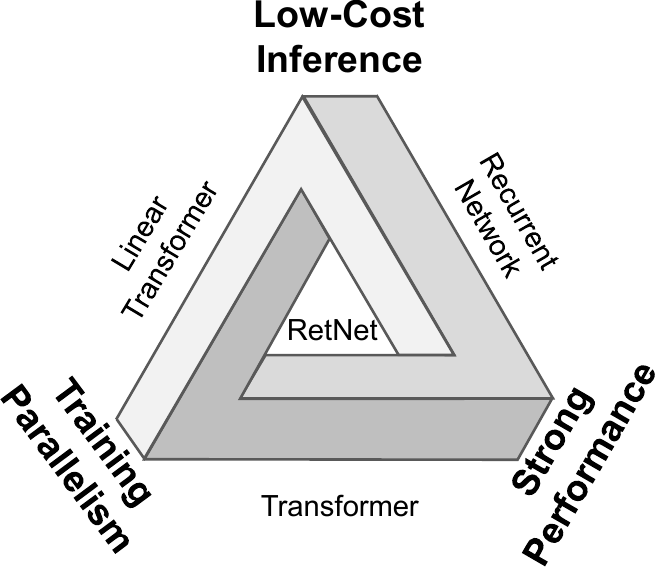}
\end{center}
\vspace{-3pt}
\caption{\our{} makes the ``impossible triangle'' possible, which achieves training parallelism, good performance, and low inference cost simultaneously.}
\vspace{-10pt}
\label{fig:triangle}
\end{wrapfigure}

Transformer~\cite{transformer} has become the de facto architecture for large language models~\cite{gpt3}, which was initially proposed to overcome the sequential training issue of recurrent models~\cite{lstm}.
However, training parallelism of Transformers is at the cost of inefficient inference, because of the $O(N)$ complexity per step and memory-bound key-value cache~\cite{multiquery}, which renders Transformers unfriendly to deployment.
The growing sequence length increases GPU memory consumption as well as latency and reduces inference speed.
 
Numerous efforts have continued to develop the next-generation architecture, aiming at retaining training parallelism and competitive performance as Transformers while having efficient $O(1)$ inference.
It is challenging to achieve the above goals simultaneously, i.e., the so-called ``impossible triangle'' as shown in \Cref{fig:triangle}.

There have been three main strands of research.
First, linearized attention~\cite{linear-transformer} approximates standard attention scores $\exp(\vq \cdot \vk)$ with kernels $\phi(\vq) \cdot \phi(\vk)$, so that autoregressive inference can be rewritten in a recurrent form.
However, the modeling capability and performance are worse than Transformers, which hinders the method's popularity.
The second strand returns to recurrent models for efficient inference while sacrificing training parallelism.
As a remedy, element-wise operators~\cite{rwkv} are used for acceleration, however, representation capacity and performance are harmed.
The third line of research explores replacing attention with other mechanisms, such as S4~\cite{s4}, and its variants~\cite{h3,hyena}.
None of the previous work can break through the impossible triangle, resulting in no clear winner compared with Transformers.

In this work, we propose retentive networks (\our{}), achieving low-cost inference, efficient long-sequence modeling, Transformer-comparable performance, and parallel model training simultaneously.
Specifically, we introduce a multi-scale retention mechanism to substitute multi-head attention, which has three computation paradigms, i.e., parallel, recurrent, and chunkwise recurrent representations.
First, the parallel representation empowers training parallelism to utilize GPU devices fully.
Second, the recurrent representation enables efficient $O(1)$ inference in terms of memory and computation. The deployment cost and latency can be significantly reduced. Moreover, the implementation is greatly simplified without key-value cache tricks.
Third, the chunkwise recurrent representation can perform efficient long-sequence modeling. We parallelly encode each local block for computation speed while recurrently encoding the global blocks to save GPU memory.

We conduct extensive experiments to compare \our{} with Transformer and its variants.
Experimental results on language modeling show that \our{} is consistently competitive in terms of both scaling curves and in-context learning.
Moreover, the inference cost of \our{} is length-invariant.
For a 7B model and 8k sequence length, \our{} decodes 8.4$\times$ faster and saves 70\% of memory than Transformers with key-value caches.
During training, \our{} also achieves 25-50\% memory saving and 7$\times$ acceleration than standard Transformer and an advantage towards highly-optimized FlashAttention~\cite{flashattention}.
Besides, \our{}'s inference latency is insensitive to batch size, allowing enormous throughput.
The intriguing properties make \our{} a strong successor to Transformer for large language models.

\vfill{}

\section{Retentive Networks}
\label{sec:retnet}

Retentive network (\our{}) is stacked with $L$ identical blocks, which follows a similar layout (i.e., residual connection, and pre-LayerNorm) as in Transformer~\cite{transformer}.
Each \our{} block contains two modules: a multi-scale retention (MSR) module, and a feed-forward network (FFN) module.
We introduce the MSR module in the following sections.
Given an input sequence $x = x_{1} \cdots x_{|x|}$, \our{} encodes the sequence in an autoregressive way.
The input vectors $\{\vx_i\}_{i=1}^{|x|}$ is first packed into $X^0 = [\vx_1, \cdots, \vx_{|x|}] \in \mathbb{R}^{|x|\times d_\text{model}}$, where $d_\text{model}$ is hidden dimension.
Then we compute contextualized vector representations $X^l = \mathrm{RetNet}_{l}(X^{l-1}), l \in [1, L]$.

\subsection{Retention}
\label{sec:retention}

In this section, we introduce the retention mechanism that has a dual form of recurrence and parallelism.
So we can train the models in a parallel way while recurrently conducting inference.

Given input $X \in \mathbb{R}^{|x|\times d_\text{model}}$, we project it to one-dimensional function $v(n) = X_n \cdot \vw_V$.
Consider a sequence modeling problem that maps $v(n) \mapsto o(n)$ through states $\vs_n$. Let $v_n, o_n$ denote $v(n), o(n)$ for simplicity.
We formulate the mapping in a recurrent manner:
\begin{equation}
\begin{aligned}
\label{eq:rnn}
&\vs_n = A\vs_{n-1} + K_n^\intercal v_n, &A\in\mathbb{R}^{d\times d}, K_n\in\mathbb{R}^{1\times d} \\
&o_n = Q_n\vs_n = \sum_{m=1}^n Q_n A^{n - m} K_m^\intercal v_m, &Q_n\in\mathbb{R}^{1\times d}
\end{aligned}
\end{equation}
where we map $v_n$ to the state vector $\vs_n$, and then implement a linear transform to encode sequence information recurrently.

Next, we make the projection $Q_n, K_n$ content-aware:
\begin{equation}
\label{eq:content:aware}
Q = X W_Q, \quad K = X W_K
\end{equation}
where $W_Q, W_K \in \mathbb{R}^{d\times d}$ are learnable matrices.

We diagonalize the matrix $A = \Lambda(\gamma e^{i \theta})\Lambda^{-1}$, where $\gamma, \theta \in \mathbb{R}^d$.
Then we obtain $A^{n - m} = \Lambda (\gamma e^{i\theta})^{n - m} \Lambda^{-1}$.
By absorbing $\Lambda$ into $W_Q$ and $W_K$, we can rewrite \Eqref{eq:rnn} as:
\begin{equation}
\begin{aligned}
\label{eq:ret:xpos}
o_n &= \sum_{m=1}^n Q_n (\gamma e^{i\theta})^{n - m}  K_m^\intercal v_m \\
&=\sum_{m=1}^n(Q_n(\gamma e^{i\theta})^n)(K_m(\gamma e^{i\theta})^{-m})^\intercal v_m
\end{aligned}
\end{equation}
where $Q_n(\gamma e^{i\theta})^n, K_m(\gamma e^{i\theta})^{-m}$ is known as xPos~\cite{lex}, i.e., a relative position embedding proposed for Transformer.
We further simplify $\gamma$ as a scalar, \Eqref{eq:ret:xpos} becomes:
\begin{equation}
\label{eq:ret:element}
o_n = \sum_{m=1}^n \gamma^{n-m} (Q_n e^{in\theta})(K_m e^{im\theta})^\dag v_m
\end{equation}
where $^\dag$ is the conjugate transpose.
The formulation is easily parallelizable within training instances.

In summary, we start with recurrent modeling as shown in \Eqref{eq:rnn}, and then derive its parallel formulation in \Eqref{eq:ret:element}.
We consider the original mapping $v(n) \mapsto o(n)$ as vectors and obtain the retention mechanism as follows.

\begin{figure*}[t]
\centering
\begin{subfigure}[b]{0.48\textwidth}
\centering
\includegraphics[width=0.5\textwidth]{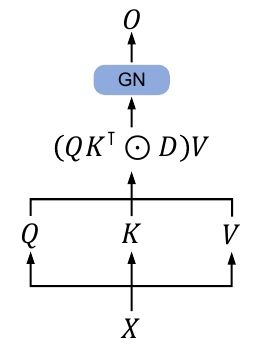}
\caption{Parallel representation.}
\label{fig:arch:parallel}
\end{subfigure}
\hfill
\begin{subfigure}[b]{0.48\textwidth}
\centering
\includegraphics[width=\textwidth]{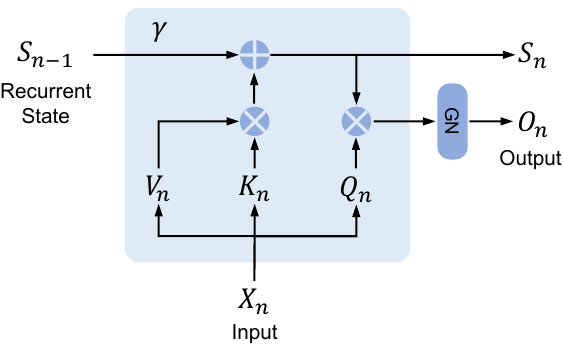}
\caption{Recurrent representation.}
\label{fig:arch:recurrent}
\end{subfigure}
\caption{Dual form of \our{}. ``GN'' is short for GroupNorm.}
\end{figure*}

\paragraph{The Parallel Representation of Retention}
As shown in \Cref{fig:arch:parallel}, the retention layer is defined as:
\begin{equation}
\begin{aligned}
\label{eq:ret:parallel}
Q = (X W_Q) \odot \Theta ,&\quad K = (X W_K) \odot \overline{\Theta} ,\quad V = X W_V \\
\Theta_n = e^{in\theta},& \quad
D_{nm} =
\left\{
\begin{aligned}
& \gamma^{n-m}, &n\ge m \\
& 0, &n < m \\
\end{aligned}
\right.
\\
\mathrm{Rete}&\mathrm{ntion}(X) = (Q K^\intercal \odot D)V
\end{aligned}
\end{equation}
where $\overline{\Theta}$ is the complex conjugate of $\Theta$, and $D \in \mathbb{R}^{|x|\times |x|}$ combines causal masking and exponential decay along relative distance as one matrix.
Similar to self-attention, the parallel representation enables us to train the models with GPUs efficiently.

\paragraph{The Recurrent Representation of Retention}
As shown in \Cref{fig:arch:recurrent}, the proposed mechanism can also be written as recurrent neural networks (RNNs), which is favorable for inference.
For the $n$-th timestep, we recurrently obtain the output as:
\begin{equation}
\begin{aligned}
\label{eq:ret:recurrent}
&S_n = \gamma S_{n-1} + K_n^{\intercal} V_n \\
&\mathrm{Rete}\mathrm{ntion} (X_n) = Q_n S_n, \quad n = 1, \cdots, |x| \\
\end{aligned}
\end{equation}
where $Q,K,V,\gamma$ are the same as in \Eqref{eq:ret:parallel}.

\paragraph{The Chunkwise Recurrent Representation of Retention}
A hybrid form of parallel representation and recurrent representation is available to accelerate training, especially for long sequences.
We divide the input sequences into chunks. Within each chunk, we follow the parallel representation (\Eqref{eq:ret:parallel}) to conduct computation. In contrast, cross-chunk information is passed following the recurrent representation (\Eqref{eq:ret:recurrent}).
Specifically, let $B$ denote the chunk length. We compute the retention output of the $i$-th chunk via:
\begin{equation}
\begin{aligned}
\label{eq:ret:chunk}
Q_{[i]} = Q_{Bi:B(i+1)} &,\quad K_{[i]} = K_{Bi:B(i+1)},\quad V_{[i]} = V_{Bi:B(i+1)} \\
R_{i}&=K_{[i]}^\intercal (V_{[i]}\odot \zeta)+\gamma^{B}R_{i-1}, \quad\zeta_{ij}=\gamma^{B-i-1}\\
\mathrm{Rete}\mathrm{ntion} (X_{[i]}) &= \underbrace{ (Q_{[i]} K^\intercal_{[i]}\odot D) V_{[i]} }_{\text{Inner-Chunk}} + \underbrace{ (Q_{[i]}R_{i-1}) \odot \xi}_{\text{Cross-Chunk}}, \quad\xi_{ij}=\gamma^{i+1}
\end{aligned}
\end{equation}
where ${[i]}$ indicates the $i$-th chunk, i.e., $x_{[i]} = [x_{(i-1)B+1} , \cdots , x_{iB}]$.

\subsection{Gated Multi-Scale Retention}
\label{sec:multiscale}

We use $h = \nicefrac{d_\text{model}}{d}$ retention heads in each layer, where $d$ is the head dimension. The heads use different parameter matrices $W_Q, W_K, W_V \in \mathbb{R}^{d\times d}$.
Moreover, \textbf{m}ulti-\textbf{s}cale \textbf{r}etention (MSR) assigns different $\gamma$ for each head. For simplicity, we set $\gamma$ identical among different layers and keep them fixed.
In addition, we add a $\mathrm{swish}$ gate~\cite{gelu,swish} to increase the non-linearity of retention layers.
Formally, given input $X$, we define the layer as:
\begin{equation}
\begin{aligned}
\label{eq:msr}
\mathbf{\gamma} &= 1 - 2^{-5-\mathrm{arange}(0, h)} \in \mathbb{R}^{h} \\
\mathrm{head}_i &= \mathrm{Retention}(X, \gamma_i) \\
Y &= \mathrm{GroupNorm}_{h}( \mathrm{Concat}(\mathrm{head}_1, \cdots, \mathrm{head}_h) ) \\
\mathrm{MSR}(X) &= (\mathrm{swish}(X W_G) \odot Y) W_O
\end{aligned}
\end{equation}
where $W_G, W_O \in \mathbb{R}^{d_\text{model}\times d_\text{model}}$ are learnable parameters, and $\mathrm{GroupNorm}$~\cite{groupnorm} normalizes the output of each head, following SubLN proposed in \cite{megatron}.
Notice that the heads use multiple $\gamma$ scales, which results in different variance statistics. So we normalize the head outputs separately.

\begin{figure}[t]
\centering
\begin{minipage}[t]{0.45\linewidth}
\centering
\begin{lstlisting}[language=python, mathescape, breaklines=true]  
def ParallelRetention(
    q, # bsz * num_head * len * qk_dim
    k, # bsz * num_head * len * qk_dim
    v, # bsz * num_head * len * v_dim
    decay_mask # num_head * len * len
    ):
    retention = q @ k.transpose(-1, -2)
    retention = retention * decay_mask
    output = retention @ v
    output = group_norm(output)
    return output
\end{lstlisting}
\end{minipage}
\begin{minipage}[t]{0.54\linewidth}
\centering
\begin{lstlisting}[language=python, mathescape, breaklines=true]
def RecurrentRetention(
    q, k, v, # bsz * num_head * len * qkv_dim
    past_kv, # bsz * num_head * qk_dim * v_dim
    decay # num_head * 1 * 1
    ):
    current_kv = decay * past_kv + k.unsqueeze(-1) * v.unsqueeze(-2)
    output = torch.sum(q.unsqueeze(-1) * current_kv, dim=-2)
    output = group_norm(output)
    return output, current_kv
\end{lstlisting}
\end{minipage}
\begin{minipage}[c]{0.8\linewidth}
\centering
\begin{lstlisting}[language=python, mathescape, breaklines=true]
def ChunkwiseRetention(
    q, k, v, # bsz * num_head * chunk_size * qkv_dim
    past_kv, # bsz * num_head * qk_dim * v_dim
    decay_mask, # num_head * chunk_size * chunk_size
    chunk_decay, # num_head * 1 * 1
    inner_decay, # num_head * chunk_size
    ):
    retention = q @ k.transpose(-1, -2)
    retention = retention * decay_mask
    inner_retention = retention @ v
    cross_retention = (q @ past_kv) * inner_decay
    retention = inner_retention + cross_retention
    output = group_norm(retention)
    current_kv = chunk_decay * past_kv + k.transpose(-1, -2) @ v
    return output, current_kv
\end{lstlisting}
\end{minipage}
\caption{Pseudocode for the three computation paradigms of retention.}
\label{fig:psuedo:code}
\end{figure}

The pseudocode of retention is summarized in \Cref{fig:psuedo:code}.

\paragraph{Retention Score Normalization}
We utilize the scale-invariant nature of $\mathrm{GroupNorm}$ to improve the numerical precision of retention layers.
Specifically, multiplying a scalar value within $\mathrm{GroupNorm}$ does not affect outputs and backward gradients, i.e., $\mathrm{GroupNorm}(\alpha * \mathrm{head}_i) = \mathrm{GroupNorm}(\mathrm{head}_i)$.
We implement three normalization factors in \Eqref{eq:ret:parallel}.
First, we normalize $Q K^\intercal$ as $\nicefrac{Q K^\intercal}{\sqrt{d}}$.
Second, we replace $D$ with $\tilde{D}_{nm} = \nicefrac{D_{nm}}{\sqrt{\sum_{i=1}^{n}D_{ni}}}$.
Third, let $R$ denote the retention scores $R=QK^\intercal \odot D$, we normalize it as $\tilde{R}_{nm} = \nicefrac{R_{nm}}{\max(| \sum_{i=1}^{n} R_{ni} |, 1)}$. Then the retention output becomes $\mathrm{Retention}(X) = \tilde{R} V$.
The above tricks do not affect the final results while stabilizing the numerical flow of both forward and backward passes, because of the scale-invariant property.

\subsection{Overall Architecture of Retention Networks}
\label{sec:arch}

For an $L$-layer retention network, we stack multi-scale retention (MSR) and feed-forward network (FFN) to build the model.
Formally, the input sequence $\{x_i\}_{i=1}^{|x|}$ is transformed to vectors by a word embedding layer. We use the packed embeddings $X^0 = [\vx_1, \cdots, \vx_{|x|}] \in \mathbb{R}^{|x|\times d_\text{model}}$ as the input and compute the model output $X^L$:
\begin{equation}
\begin{aligned}
\label{eq:arch}
Y^l &= \mathrm{MSR}(\mathrm{LN}(X^l)) + X^l \\
X^{l+1} &= \mathrm{FFN}(\mathrm{LN}(Y^l)) + Y^l
\end{aligned}
\end{equation}
where $\mathrm{LN}(\cdot)$ is LayerNorm~\cite{layernorm}.
The FFN part is computed as $\mathrm{FFN}(X) = \mathrm{gelu}(X W_1) W_2$, where $W_1, W_2$ are parameter matrices.

\paragraph{Training}
We use the parallel (\Eqref{eq:ret:parallel}) and chunkwise recurrent 
(\Eqref{eq:ret:chunk}) representations during the training process.
The parallelization within sequences or chunks efficiently utilizes GPUs to accelerate computation.
More favorably, chunkwise recurrence is especially useful for long-sequence training, which is efficient in terms of both FLOPs and memory consumption.

\paragraph{Inference}
The recurrent representation (\Eqref{eq:ret:recurrent}) is employed during the inference, which nicely fits autoregressive decoding.
The $O(1)$ complexity reduces memory and inference latency while achieving equivalent results.

\subsection{Relation to and Differences from Previous Methods}

Table~\ref{tbl:compare:aspects} compares \our{} with previous methods from various perspectives.
The comparison results echo the ``impossible triangle'' presented in Figure~\ref{fig:triangle}.
Moreover, \our{} has linear memory complexity for long sequences due to the chunkwise recurrent representation.
We also summarize the comparisons with specific methods as follows.

\paragraph{Transformer}
The parallel representation of retention shares similar spirits as Transformers~\cite{transformer}.
The most related Transformer variant is Lex Transformer~\cite{lex} which implements xPos as position embeddings.
As described in \Eqref{eq:ret:xpos}, the derivation of retention aligns with xPos.
In comparison with attention, retention removes $\softmax$ and enables recurrent formulation, which significantly benefits inference.

\paragraph{S4}
Unlike \Eqref{eq:content:aware}, if $Q_n$ and $K_n$ are content-unaware, the formulation can be degenerated to S4~\cite{s4}, where $O = (Q K^\intercal, Q A K^\intercal, .., Q A^{|x|-1} K^\intercal) * V$.

\paragraph{Linear Attention}
The variants typically use various kernels $\nicefrac{\phi(q_i)\phi(k_j)}{\sum_{n=1}^{|x|}\phi(q_i)\phi(k_n)}$ to replace the $\softmax$ function.
However, linear attention struggles to effectively encode position information, rendering the models less performant.
Besides, we reexamine sequence modeling from scratch, rather than aiming at approximating $\softmax$.

\paragraph{AFT/RWKV}
Attention Free Transformer (AFT) simplifies dot-product attention to element-wise operations and moves $\softmax$ to key vectors.
RWKV replaces AFT's position embeddings with exponential decay and runs the models recurrently for training and inference.
In comparison, retention preserves high-dimensional states to encode sequence information, which contributes to expressive ability and better performance.

\paragraph{xPos/RoPE}
Compared with relative position embedding methods proposed for Transformers, \Cref{eq:ret:xpos} presents a similar formulation as xPos~\cite{lex} and RoPE~\cite{rotary}.

\paragraph{Sub-LayerNorm}
As shown in \Eqref{eq:msr}, the retention layer uses Sub-LayerNorm~\cite{magneto} to normalize outputs. Because the multi-scale modeling leads to different variances for the heads, we replace the original LayerNorm with GroupNorm.

\begin{table*}[t]
\begin{tabular}{lcccc}
\toprule
\bf Architectures      & \bf \tabincell{c}{Training \\ Parallelization} & \bf  Inference Cost & \bf \tabincell{c}{Long-Sequence \\ Memory Complexity} & \bf  Performance \\
\midrule
Transformer        & \ding{52} & $O(N)$ & $O(N^2)$ & \ding{52}\ding{52} \\
Linear Transformer & \ding{52} & $O(1)$ & $O(N)$ & \ding{56} \\
Recurrent NN       & \ding{56} & $O(1)$ & $O(N)$ & \ding{56} \\
RWKV               & \ding{56} & $O(1)$ & $O(N)$ & \ding{52} \\
H3/S4              & \ding{52} & $O(1)$ & $O(N \log N)$ & \ding{52} \\
Hyena              & \ding{52} & $O(N)$ & $O(N\log N)$ & \ding{52} \\
\our{}                & \ding{52} & $O(1)$ & $O(N)$ & \ding{52}\ding{52} \\
\bottomrule
\end{tabular}
\caption{Model comparison from various perspectives. \our{} achieves training parallelization, constant inference cost, linear long-sequence memory complexity, and good performance.}
\label{tbl:compare:aspects}
\end{table*}

\section{Experiments}
\label{sec:exp}

We conduct experiments on language modeling to evaluate \our{}.
We evaluate the proposed architecture with various benchmarks, i.e., language modeling performance, and zero-/few-shot learning on downstream tasks.
Moreover, for training and inference, we compare speed, memory consumption, and latency.

\subsection{Setup}
\label{sec:setup}

\paragraph{Parameter Allocation}
We re-allocate the parameters in MSR and FFN for fair comparisons.
Let $d$ denote $d_\text{model}$ for simplicity here.
In Transformers, there are about $4d^2$ parameters in self-attention where $W_Q, W_K, W_V, W_O\in\mathbb{R}^{d\times d}$, and $8d^2$ parameters in FFN where the intermediate dimension is $4d$.
In comparison, \our{} has $8d^2$ parameters in retention, where $W_Q, W_K\in\mathbb{R}^{d\times d}, W_G,W_V\in\mathbb{R}^{d\times 2d}, W_O\in\mathbb{R}^{2d\times d}$.
Notice that the head dimension of $V$ is twice $Q, K$. The widened dimension is projected back to $d$ by $W_O$.
In order to keep the parameter number the same as Transformer, the FFN intermediate dimension in \our{} is $2d$.
Meanwhile, we set the head dimension to $256$ in our experiments, i.e., $256$ for queries and keys, and $512$ for values.
For fair comparison, we keep $\mathbf{\gamma}$ identical among different model sizes, where $\mathbf{\gamma} = 1 - e^{\mathrm{linspace}(\log\nicefrac{1}{32}, \log\nicefrac{1}{512}, h)} \in \mathbb{R}^{h}$ instead of the default value in \Eqref{eq:msr}.

\begin{table*}[t]
\centering
\captionsetup{type=table}
\begin{tabular}{cccccc}
\toprule
\bf Size & \bf Hidden Dim. & \bf \#Layers & \bf Batch Size & \bf \# Tokens & \bf Learning Rate \\
\midrule
1.3B & 2048 & 24 & 4M & 100B & $6\times10^{-4}$ \\
2.7B & 2560 & 32 & 4M & 100B & $3\times10^{-4}$ \\
6.7B & 4096 & 32 & 4M & 100B & $3\times10^{-4}$ \\
\bottomrule
\end{tabular}
\caption{Sizes, and learning hyper-parameters of the models in language modeling experiments.}
\label{tbl:size}
\end{table*}

\paragraph{Language Model Training}
As shown in Table~\ref{tbl:size}, we train language models with various sizes (i.e., 1.3B, 2.7B, and 6.7B) from scratch.
The training corpus is a curated compilation of The Pile~\cite{pile}, C4~\cite{c4:ai2}, and The Stack~\cite{TheStack}.
We append the \texttt{<bos>} token to indicate the start of a sequence\footnote{We find that appending the \texttt{<bos>} token at the beginning benefits training stability and performance.}.
The training batch size is 4M tokens with 2048 maximal length.
We train the models with 100B tokens, i.e., 25k steps.
We use the AdamW~\cite{adamw} optimizer with $\beta_1=0.9, \beta_2=0.98$, and weight decay is set to $0.05$.
The number of warmup steps is 375 with linear learning rate decay.
The parameters are initialized following DeepNet~\cite{deepnet} to guarantee training stability.
The implementation is based on TorchScale~\cite{torchscale}.
We train the models with 512 AMD MI200 GPUs.

\subsection{Comparisons with Transformer}

\begin{figure*}[t]
\centering
\centering
\captionsetup{type=figure}
\includegraphics[width=0.5\textwidth]{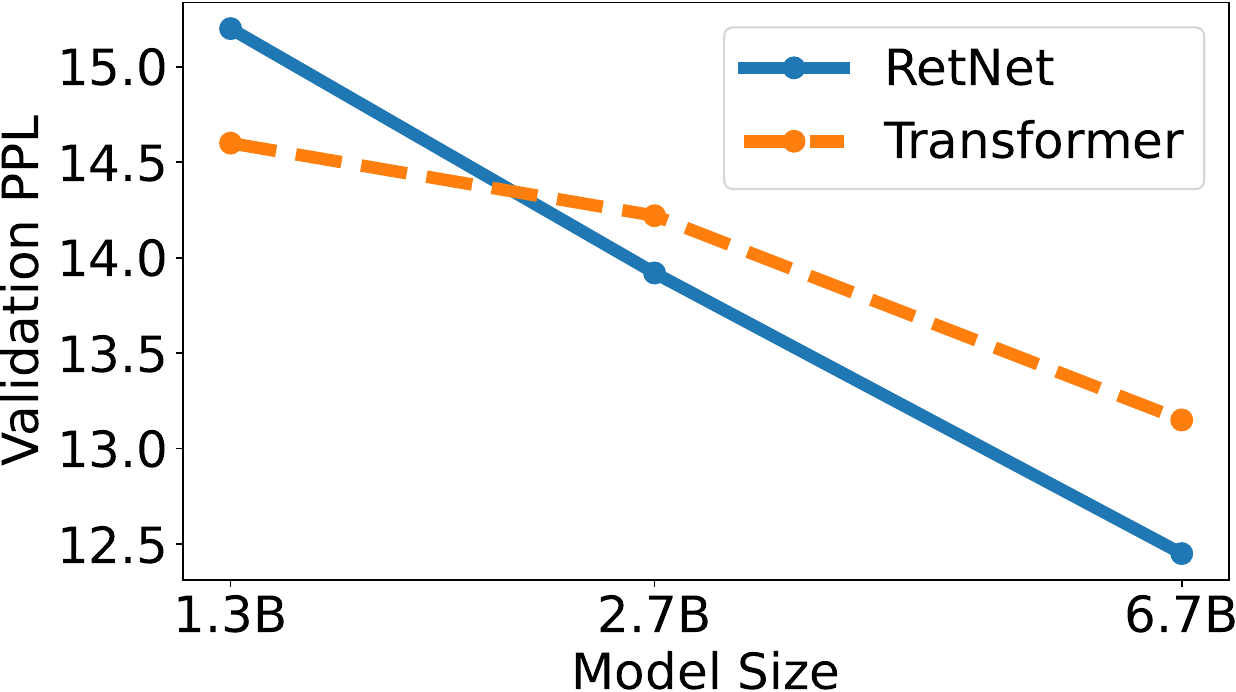}
\caption{Perplexity decreases along with scaling up the model size. We empirically observe that \our{} tends to outperform Transformer when the model size is larger than 2B.}
\label{fig:ppl}
\end{figure*}

\paragraph{Language Modeling}
As shown in Figure~\ref{fig:ppl}, we report perplexity on the validation set for the language models based on Transformer and \our{}.
We present the scaling curves with three model sizes, i.e., 1.3B, 2.7B, and 6.7B.
\our{} achieves comparable results with Transformers.
More importantly, the results indicate that \our{} is favorable regarding size scaling.
Besides performance, the \our{} training is quite stable in our experiments.
Experimental results show that \our{} is a strong competitor to Transformer for large language models.
Empirically, we find that \our{} starts to outperform Transformer when the model size is larger than 2B.
We also summarize the language modeling results with different context lengths in Appendix~\ref{app:ppl:length}.

\begin{table*}
\centering
\begin{tabular}{@{}lcccccccc@{}}
\toprule
& \textbf{HS} & \textbf{BoolQ} & \textbf{COPA} & \textbf{PIQA} & \textbf{Winograd} & \textbf{Winogrande} & \textbf{SC} & \bf Avg \\
\midrule
\multicolumn{8}{l}{\textit{Zero-Shot}} & \\
Transformer & 55.9 & 62.0 & 69.0 & 74.6 & 69.5 & 56.5 & 75.0 & 66.07\\
\our{} & \textbf{60.7} & \textbf{62.2} & \textbf{77.0} & \textbf{75.4} & \textbf{77.2} & \textbf{58.1} & \textbf{76.0} & \bf 69.51 \\
\midrule
\multicolumn{8}{l}{\textit{4-Shot}} & \\
Transformer & 55.8 & 58.7 & 71.0 & 75.0 & 71.9 & 57.3 & 75.4 & 66.44\\
\our{} & \textbf{60.5} & \textbf{60.1} & \textbf{78.0} & \textbf{76.0} & \textbf{77.9} & \textbf{59.9} & \textbf{75.9} & \bf 69.76 \\
\bottomrule
\end{tabular}
\caption{Zero-shot and few-shot learning with Transformer and \our{}. The model size is 6.7B.}
\label{tbl:endtask}
\end{table*}

\paragraph{Zero-Shot and Few-Shot Evaluation on Downstream Tasks}
We also compare the language models on a wide range of downstream tasks.
We evaluate zero-shot and 4-shot learning with the 6.7B models.
As shown in Table~\ref{tbl:endtask}, the datasets include HellaSwag (HS)~\cite{hellaswag}, BoolQ~\cite{boolq}, COPA~\cite{superglue}, PIQA~\cite{piqa}, Winograd, Winogrande~\cite{winograd2012}, and StoryCloze (SC)~\cite{storycloze}.
The accuracy numbers are consistent with language modeling perplexity presented in Figure~\ref{fig:ppl}.
\our{} achieves comparable performance with Transformer on zero-shot and in-context learning settings.

\subsection{Training Cost}
\label{sec:training:cost}

\begin{table*}[t]
\centering
\begin{tabular}{l ccc ccc}
\toprule
\multirow{2}{*}{\textbf{Model Size}} & \multicolumn{3}{c|}{\bf Memory (GB) $\downarrow$} & \multicolumn{3}{c}{\bf Throughput (wps) $\uparrow$} \\
 & Trm & Trm+FlashAttn & \our{} & Trm & Trm+FlashAttn & \our{} \\
\midrule
1.3B & 74.8 & 38.8 & 34.5 & 10832.4 & 63965.2 & 73344.8 \\
2.7B & 69.6 & 42.1 & 42.0 & 5186.0 & 34990.2 & 38921.2 \\
6.7B & 69.0 & 51.4 & 48.0 & 2754.4 & 16230.1 & 17458.6 \\
13B  & 61.4 & 46.3 & 45.9 & 1208.9 & 7945.1 & 8642.2 \\
\bottomrule
\end{tabular}
\caption{Training cost of Transformer (Trm), Transformer with FlashAttention (Trm+FlashAttn), and \our{}. We report memory consumption and training throughput (word per second; wps).}
\label{tbl:traincost}
\end{table*}

As shown in Table~\ref{tbl:traincost}, we compare the training speed and memory consumption of Transformer and \our{}, where the training sequence length is 8192.
We also compare with FlashAttention~\cite{flashattention}, which improves speed and reduces GPU memory IO by recomputation and kernel fusion.
In comparison, we implement \our{} using vanilla PyTorch code, and leave kernel fusion or FlashAttention-like acceleration for future work.
We use chunkwise recurrent representation of retention as described in \Eqref{eq:ret:chunk}. The chunk size is set to $512$.
We evaluate the results with eight Nvidia A100-80GB GPUs, because FlashAttention is highly optimized for A100.
Tensor parallelism is enabled for 6.7B and 13B models.

Experimental results show that \our{} is more memory-efficient and has higher throughput than Transformers during training.
Even compared with FlashAttention, \our{} is still competitive in terms of speed and memory cost.
Moreover, without relying on specific kernels, it is easy to train \our{} on other platforms efficiently. For example, we train the \our{} models on an AMD MI200 cluster with decent throughput.
It is notable that \our{} has the potential to further reduce cost via advanced implementation, such as kernel fusion.

\subsection{Inference Cost}
\label{sec:inference:cost}

\begin{figure*}[ht]
\centering
\begin{subfigure}[b]{0.48\textwidth}
\centering
\includegraphics[width=\textwidth]{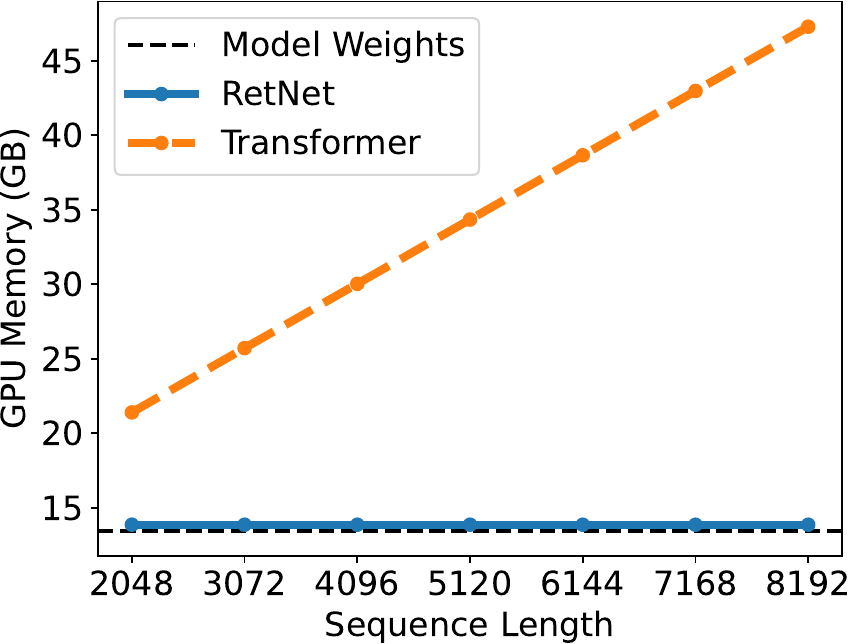}
\caption{GPU memory cost of Transformer and \our{}.}
\label{fig:memorycost}
\end{subfigure}
\hfill
\begin{subfigure}[b]{0.48\textwidth}
\centering
\includegraphics[width=\textwidth]{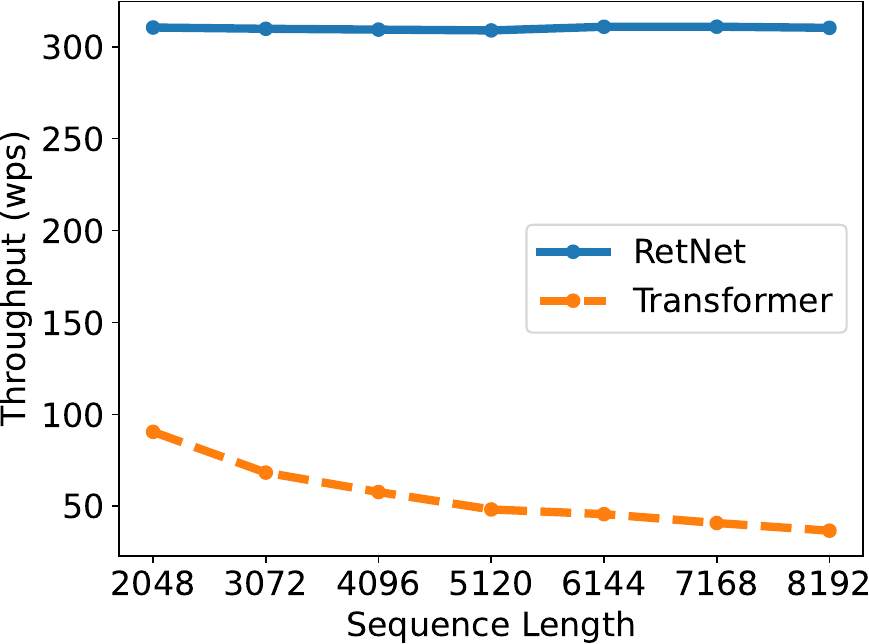}
\caption{Throughput of Transformer and \our{}.}
\label{fig:speedcost}
\end{subfigure}
\hfill
\begin{subfigure}[b]{0.48\textwidth}
\centering
\includegraphics[width=\textwidth]{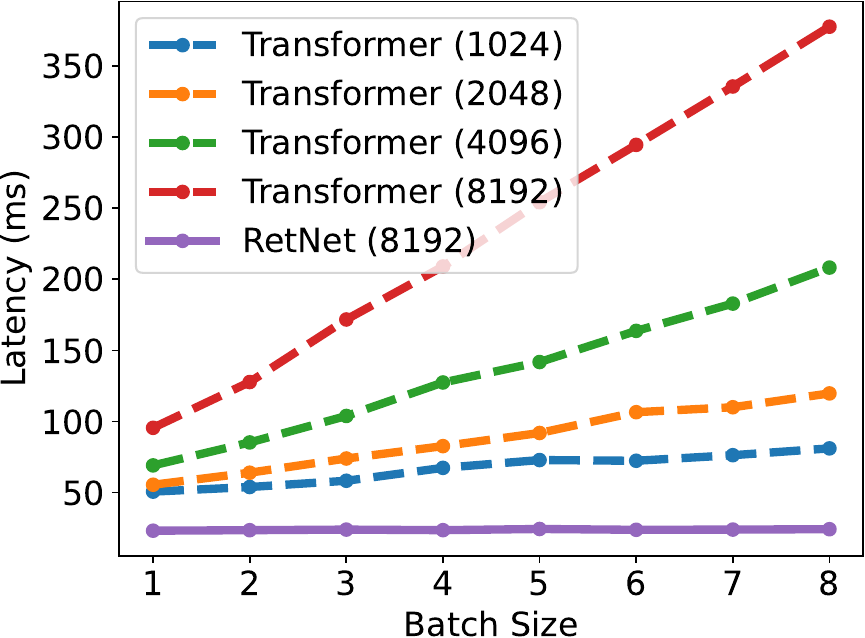}
\caption{Inference latency with different batch sizes.}
\label{fig:latency}
\end{subfigure}
\caption{Inference cost of Transformer and \our{} with a model size of 6.7B. \our{} outperforms Transformers in terms of memory consumption, throughput, and latency.}
\label{fig:infer:cost}
\end{figure*}

As shown in Figure~\ref{fig:infer:cost}, we compare memory cost, throughput, and latency of Transformer and \our{} during inference.
Transformers reuse KV caches of previously decoded tokens.
\our{} uses the recurrent representation as described in \Eqref{eq:ret:recurrent}.
We evaluate the 6.7B model on the A100-80GB GPU in our experiments.
Figure~\ref{fig:infer:cost} shows that \our{} outperforms Transformer in terms of inference cost.

\paragraph{Memory}
As shown in Figure~\ref{fig:memorycost}, the memory cost of Transformer increases linearly due to KV caches.
In contrast, the memory consumption of \our{} remains consistent even for long sequences, requiring much less GPU memory to host \our{}.
The additional memory consumption of \our{} is almost negligible (i.e., about 3\%) while the model weights occupy 97\%.

\paragraph{Throughput}
As presented in Figure~\ref{fig:speedcost}, the throughput of Transformer drops along with the decoding length increases. In comparison, \our{} has higher and length-invariant throughput during decoding, by utilizing the recurrent representation of retention.

\paragraph{Latency}
Latency is an important metric in deployment, which greatly affects user experience.
We report decoding latency in Figure~\ref{fig:latency}.
Experimental results show that increasing batch size renders Transformer's latency larger.
Moreover, the latency of Transformers grows faster with longer input.
In order to make latency acceptable, we have to restrict the batch size, which harms the overall inference throughput of Transformers.
By contrast, \our{}'s decoding latency outperforms Transformers and keeps almost the same across different batch sizes and input lengths.

\subsection{Comparison with Transformer Variants}
\label{sec:competitor}

\begin{table*}[t]
\centering
\begin{tabular}{lc|cccc}
\toprule
\textbf{Method}    & \textbf{In-Domain}      &  \textbf{PG22}   &   \textbf{QMSum}    & \textbf{GovReport} & \textbf{SummScreen}\\
\midrule
RWKV & 30.92 & 51.41 & 28.17 & 19.80 & 25.78 \\
H3 & 29.97 & 49.17 & 24.29 & 19.19 & 25.11 \\
Hyena & 32.08 & 52.75 & 28.18 & 20.55 & 26.51 \\
Linear Transformer & 40.24 & 63.86 & 28.45 & 25.33 & 32.02 \\
\our{} & \textbf{26.05} & \textbf{45.27} & \textbf{21.33} & \textbf{16.52} & \textbf{22.48} \\
\bottomrule
\end{tabular}
\caption{Perplexity results on language modeling. \our{} outperforms other architectures on both the in-domain evaluation set and various out-of-domain corpora.}
\label{tbl:competitor}
\end{table*}

Apart from Transformer, we compare \our{} with various efficient Transformer variants, including Linear Transformer~\cite{linear-transformer}, RWKV~\cite{rwkv}, H3~\cite{h3}, and Hyena~\cite{hyena}.
All models have 200M parameters with 16 layers and a hidden dimension of 1024.
For H3, we set the head dimension as 8.
For RWKV, we use the TimeMix module to substitute self-attention layers while keeping FFN layers consistent with other models for fair comparisons.
We train the models with 10k steps with a batch size of 0.5M tokens.
Most hyperparameters and training corpora are kept the same as in Section~\ref{sec:setup}.

Table~\ref{tbl:competitor} reports the perplexity numbers on the in-domain validation set and other out-of-domain corpora, e.g., Project Gutenberg 2019-2022 (PG22)~\cite{lex}, QMSum~\cite{qmsum}, GovReport~\cite{govreport}, SummScreen~\cite{summscreen,scrolls}.
Overall, \our{} outperforms previous methods across different datasets.
\our{} not only achieves better evaluation results on the in-domain corpus but also obtains lower perplexity on several out-of-domain datasets.
The favorable performance makes \our{} a strong successor to Transformer, besides the benefits of significant cost reduction (\Cref{sec:training:cost,sec:inference:cost}).

In addition, we discuss the training and inference efficiency of the compared methods.
Let $d$ denote the hidden dimension, and $n$ the sequence length.
For training, RWKV's token-mixing complexity is $O(dn)$ while Hyena's is $O(dn\log n)$ with Fast Fourier Transform acceleration.
The above two methods reduce training FLOPS via employing element-wise operators to trade-off modeling capacity.
In comparison with retention, the chunk-wise recurrent representation is $O(dn(b+h))$, where $b$ is the chunk size, $h$ is the head dimension, and we usually set $b=512, h=256$.
For either large model size (i.e., larger $d$) or sequence length, the additional $b+h$ has negligible effects.
So the \our{} training is quite efficient without sacrificing the modeling performance.
For inference, among the compared efficient architectures, Hyena has the same complexity (i.e., $O(n)$ per step) as Transformer while the others can perform $O(1)$ decoding.

\subsection{Ablation Studies}
\label{sec:ablation}

\begin{table*}[t]
\centering
\begin{tabular}{lc|cccc}
\toprule
\textbf{Method}    & \textbf{In-Domain}      &  \textbf{PG22}   &   \textbf{QMSum}    & \textbf{GovReport} & \textbf{SummScreen}\\
\midrule
\our{} & \textbf{26.05} & \textbf{45.27} & \textbf{21.33} & \textbf{16.52} & \textbf{22.48} \\
~~$-$ $\mathrm{swish}$ gate & 27.84 & 49.44 & 22.52 & 17.45 & 23.72 \\
~~$-$ $\mathrm{GroupNorm}$ & 27.54 & 46.95 & 22.61 & 17.59 & 23.73 \\
~~$-$ $\gamma$ decay & 27.86 & 47.85 & 21.99 & 17.49 & 23.70 \\
~~$-$ multi-scale decay & 27.02 & 47.18 & 22.08 & 17.17 & 23.38 \\
~~Reduce head dimension & 27.68 & 47.72 & 23.09 & 17.46 & 23.41 \\
\bottomrule
\end{tabular}
\caption{Ablation results on in-domain and out-of-domain corpora.}
\label{tbl:ablation}
\end{table*}

We ablate various design choices of \our{} and report the language modeling results in \Cref{tbl:ablation}.
The evaluation settings and metrics are the same as in \Cref{sec:competitor}.

\paragraph{Architecture}
We ablate the $\mathrm{swish}$ gate and $\mathrm{GroupNorm}$ as described in \Cref{eq:msr}.
\Cref{tbl:ablation} shows that the above two components improve the final performance.
Firstly, the gating module is essential for enhancing non-linearity and improving model capability. Notice that we use the same parameter allocation as Transformers after removing the gate.
Secondly, group normalization in retention balances the variances of multi-head outputs, which improves training stability and language modeling results.

\paragraph{Multi-Scale Decay}
\Cref{eq:msr} shows that we use different $\mathbf{\gamma}$ as the decay rates for the retention heads.
In the ablation studies, we examine removing $\gamma$ decay (i.e., ``$-$ $\gamma$ decay'') and applying the same decay rate across heads (i.e., ``$-$ multi-scale decay'').
Specifically, ablating $\gamma$ decay is equivalent to $\gamma = 1$.
In the second setting, we set $\gamma = 127/128$ for all heads.
\Cref{tbl:ablation} indicates that both the decay mechanism and using multiple decay rates can improve the language modeling performance.

\paragraph{Head Dimension}
From the recurrent perspective of \Cref{eq:rnn}, the head dimension implies the memory capacity of hidden states.
In the ablation study, we reduce the default head dimension from $256$ to $64$, i.e., $64$ for queries and keys, and $128$ for values.
We keep the hidden dimension $d_\text{model}$ the same so the number of heads increases.
Experimental results in \Cref{tbl:ablation} show that the larger head dimension achieves better performance.

\section{Conclusion}

In this work, we propose retentive networks (\our{}) for sequence modeling, which enables various representations, i.e., parallel, recurrent, and chunkwise recurrent.
\our{} achieves significantly better inference efficiency (in terms of memory, speed, and latency), favorable training parallelization, and competitive performance compared with Transformers.
The above advantages make \our{} an ideal successor to Transformers for large language models, especially considering the deployment benefits brought by the $O(1)$ inference complexity.
In the future, we would like to scale up \our{} in terms of model size~\cite{xmoe} and training steps.
Moreover, retention can efficiently work with structured prompting~\cite{structured:prompt} by compressing long-term memory.
We will also use \our{} as the backbone architecture to train multimodal large language models~\cite{metalm,kosmos-1,kosmos-2}.
In addition, we are interested in deploying \our{} models on various edge devices, such as mobile phones.

\section*{Acknowledgement}

We would like to acknowledge Jiayu Ding, Songlin Yang, and colleagues from MSRA System Group for the helpful discussions.

\nocite{s4}
\nocite{ResurrectingRNN}
\nocite{rwkv}

\bibliographystyle{alpha}
\bibliography{arch,anthology}

\newpage
\appendix

\section{Hyperparameters}
\label{app:hp}

\begin{table}[ht]
\centering
\begin{tabular}{lccc}
\toprule
\bf Hyperparameters & \bf 1.3B & \bf 2.7B & \bf 6.7B \\
\midrule
Layers & 24 & 32 & 32 \\
Hidden size & 2048 & 2560 & 4096 \\
FFN size & 4096 & 5120 & 8192 \\
Heads & 8 & 10 & 16 \\
\midrule
Learning rate & $6\times10^{-4}$ & $3\times10^{-4}$ & $3\times10^{-4}$ \\
LR scheduler & \multicolumn{3}{c}{Polynomial decay} \\
Warm-up steps & \multicolumn{3}{c}{375} \\
Tokens per batch & \multicolumn{3}{c}{4M} \\
Adam $\beta$ & \multicolumn{3}{c}{(0.9, 0.98)} \\
Training steps & \multicolumn{3}{c}{25,000} \\
\midrule
Gradient clipping & \multicolumn{3}{c}{2.0} \\
Dropout & \multicolumn{3}{c}{0.1} \\
Weight decay & \multicolumn{3}{c}{0.01} \\
\bottomrule
\\
\end{tabular}
\caption{Hyperparamters used for the models in~\Cref{sec:exp}.
}
\end{table}

\section{Grouped Results of Different Context Lengths}
\label{app:ppl:length}

As shown in \Cref{tbl:ppl:length}, we report language modeling results with different context lengths.
In order to make the numbers comparable, we use 2048 text chunks as evaluation data and only compute perplexity for the last 128 tokens.
Experimental results show that \our{} outperforms Transformer across different context lengths.
Besides, \our{} can utilize longer context for better results.

\begin{table*}[ht]
\centering
\begin{tabular}{lccc}
\toprule
\bf Model & \bf 512 & \bf 1024 & \bf 2048 \\
\midrule
Transformer & 13.55 & 12.56 & 12.35 \\
\our{} & 13.09 & 12.14 & 11.98 \\
\bottomrule
\end{tabular}
\caption{Language modeling perplexity of \our{} and Transformer with different context length.
The results show that \our{} has a consistent advantage across sequence length.
}
\label{tbl:ppl:length}
\end{table*}


\end{document}